\documentclass[twoside,leqno,twocolumn]{article}
\usepackage{ltexpprt}

\usepackage{url}
\usepackage{wrapfig}
\usepackage[algo2e,ruled]{algorithm2e}
\usepackage{graphicx} 
\usepackage{subfigure}
\usepackage{multirow,float}
\usepackage{xspace}
\usepackage{subfigure}
\usepackage{amsmath}
\usepackage{steinmetz}

\usepackage{amssymb}
\usepackage{graphicx}
\usepackage[toc,page]{appendix}
\usepackage{authblk}

\usepackage{url}
\usepackage{subfigure}

\usepackage{steinmetz}
\usepackage{lipsum}
\usepackage[margin=1in]{geometry}
\usepackage{amsmath}

\begin{document}

\title{Efficient Estimation of Generalization Error and Bias-Variance Components of Ensembles}


\author{Dhruv Mahajan\footnote{Facebook, USA, email: dhruv.mahajan@gmail.com (past affiliation during work : Microsoft Corporation, USA )} , Vivek Gupta\footnote{Microsoft Research, Bangalore, India, email: t-vigu@microsoft.com (research intern during work)}, S Sathiya Keerthi \footnote{Microsoft Corporation, USA, email:keerthi@microsoft.com } , Sellamanickam Sundararajan\footnote{Microsoft Research, Bangalore, India, email: ssrajan@microsoft.com} , Shravan Narayanamurthy\footnote{Microsoft Corporation, USA, email: shravan@microsoft.com} , Rahul Kidambi$^*$ \footnote{* University of Washington, email: rkidambi@uw.edu}}


%

\date{}
\maketitle

\def\prb{p_{r_k}}
\def\prbp{p_{r_k}^+}
\def\crb{c_{r_k}}
\def\Ktilde{\tilde{K}}

\begin{abstract}
For many applications, an ensemble of base classifiers is an effective solution. The tuning of its parameters (number of classifiers, amount of data on which each classifier is to be trained on, etc.) requires $G$, the generalization error of a given ensemble. The efficient estimation of $G$ is the focus of this paper. The key idea is to approximate the variance of the class scores/probabilities of the base classifiers over the randomness imposed by the training subset by normal/beta distribution at each point $x$ in the input feature space. We estimate the parameters of the distribution using a small set of randomly chosen base classifiers and use those parameters to give efficient estimation schemes for $G$. We give empirical evidence for the quality of the various estimators. We also demonstrate their usefulness in making design choices such as the number of classifiers in the ensemble and the size of subset of data used for training that are needed to achieve a certain value of generalization error. Our approach also has great potential for designing distributed ensemble classifiers.



\end{abstract}







\def\prb{p_{r_k}}
\def\prbp{p_{r_k}^+}
\def\crb{c_{r_k}}
\def\Ktilde{\tilde{K}}

\section{Introduction}
\label{sec:intro}

Ensembles of classifiers randomly picked from a collection of base classifiers are well-known to improve over the individual base classifiers. Apart from providing improved performance (e.g., accuracy), building an ensemble of classifiers in the distributed setting is a useful approach for large scale applications due to the ease of training and testing. While there are many ways of forming ensembles, we consider ensembles formed via pasting small bites\footnote{A {\it bite} refers to a subset of the full data. By a {\it small bite} we do not mean that the amount of data in the bite is small or only subset of features is used; it just means that the fraction of the full data with full features that is in the bite is small. } - a variation of bagging in which the base classifiers are trained on small bites (subsets) of a given labeled dataset~\cite{Breiman:96}. The value of such ensembles has been demonstrated in several papers;~\cite{Mokeddem:09} gives a survey of several methods in this area. Note that this way of forming ensembles is quite appropriate in the distributed setting, as the bites are formed in each node and models can be built cheaply in an embarrassingly parallel fashion. To achieve good generalization performance, designing ensemble classifiers involves making several choices: the optimal number of classifiers/models, the mixing scheme to be used, the optimal size of training sample for each base classifier etc. A key question then is, \textit{how do we estimate the generalization performance of ensembles and that too efficiently, so as to make practical choices?} Estimation of generalization error involves considering the randomness involved in the bites used for learning the models and evaluating their performance over possible inputs and outputs. This needs to be done for different parameter choices. Our approach involves the modeling, for each data point ($x$) in the input space, of the distribution of either the score or the class probability of classifier models\footnote{The classifier models could belong to any one type, e.g., linear classifiers, kernel machines , decision trees, etc.} for the randomness involved (e.g., bite selection). 




There has been a lot of work in the literature~\cite{Friedman:97,Schapire:97,Zhang:13,Christmann:07} on analyzing the generalization error of ensembles. Although, these works help in developing the theoretical understanding, those analysis do not lead to useful and efficient procedures for the estimation of either generalization error or bias-variance. Motivated by the above observations, the goal of this paper is to address the important problem of analytically estimating the generalization error and bias-variance for ensembles; and, show how these analytical estimates can be efficiently computed and utilized to make the important design choices discussed above. 


Using these distributions, we can compute the generalization error by taking the expectation of a loss function (e.g., 0-1 loss, squared error or logistic loss) defined between the model predicted and true class distributions. Following are the main contributions of this paper.


\begin{enumerate}
    \setlength\itemsep{0.2em}
    \item The key idea of modeling point-wise variations of posterior class probability via a suitable distribution, leading to efficient computations of estimates of expected generalization performance, bias and variance of ensemble classifiers, is novel.
    \item Our experiments indicate practical utility of the estimators in choosing ensemble size ($K$) by building just a few (say 25) base classifiers and universal lookup table (more efficient) . Complex data mining problems require hundreds or thousands of classifiers to do well; naturally, therefore, we expect our method to be useful for reducing computational costs. Moreover, our analysis also allows us to determine which mixing strategy is better: voting or parameter mixing.
    \item For the least squares setting we derive analytical formulas for bias-variance components for voting and parameter mixing, which is of separate interest for deriving analytical insights.
\end{enumerate}  

In Section \ref{sec:mathformal}, we introduce mathematical notation needed to explain our ideas more formally. Section \ref{sec:rw} covers related  works in the literature. Section \ref{sec:maincomp} and \ref{sec:bv} discuss the main idea of estimation of generalization error, bias and variance components. Usefulness of our analytical results and efficient estimation procedure are demonstrated through many experiments in Section \ref{sec:expts}.

\section{Mathematical Formalization}
\label{sec:mathformal}
Let $r$ denote the randomness associated with forming an ensemble formed using small bites. Suppose we are given a dataset, $D$ having $n$ labeled examples. The $k$-th small bite is a subset of size, $m < n$, chosen randomly (with replacement) from $D$; let $r_k$ denote this randomness associated with classifier from this sample bite, and $r=\{r_k\}$. Let us say that we work with one method of generating base classifiers, e.g., a kernel method, a decision tree, etc., so that the $r_k$ can be assumed to be {\em iid}.  We form the $k$-th base classifier by training on the $k$-th small bite. We will assume that, for each input $x$, these classifiers output a set of posterior class probability vectors, $\{\prb(x)\}$. The posterior class probability vector of the ensemble, $p_r(x)$ can be formed by using some mixing method on either the probability vectors $\prb(x)$ or the base classifier scores that lead to those probability vectors. Voting and parameter mixing~\cite{Mann:09,zinkevich:10} are two mixing schemes that we consider in this paper.

Let $p^*(x)$ be the true class probability vector at a given $x$. 
To make design choices such as $m$, $K$ and the mixing method, it is important to estimate $G$ i.e.
\begin{equation*}
G=E_r E_x L(p^*(x), p_r(x))
\label{eq:morig}
\end{equation*}
for each set of choices of those parameters. In generalization studies it is usual to include another level of expectation - over the choice of the training set. We do not do this here for the sake of simplicity. If the given dataset is large or $m\ll n$, then that extra level of expectation is not important. From a practical point of view, the expectation, $E_r$ in (\ref{eq:morig}) is crucial because designers would like to know the expectation and variance of $E_x L(p^*(x), p_r(x))$ over $r$ in order to make cleaner evaluations and hyper-parameter choices. 


A simple baseline approach to estimate $G$ is to use direct sampling for a range of values of $K$, the number of classifier models, and choose $K_{\min}$ to be the $K$ that yields the smallest generalization error estimate. In this approach, we form a set of $\Ktilde$ base classifiers, form several ensembles of size $K$ using it, and obtain the generalization error as the mean of the error computed on these instances of ensembles. It is important that $\Ktilde$ is decently larger than the largest $K$ value of interest; and so, this baseline method can be quite expensive.  Therefore, it is useful and natural to ask the following question: {\it By forming only a small random set of base classifier instances, say 25 of them, is it possible to estimate $G$ efficiently for all values of $K$?}. We give an approach that positively answers this question. Clearly, such an approach has implications for efficient ensemble design.



Rewrite $G$ as $G = E_x (E_r L(p^*(x), p_r(x)))$. For each base classifier (with randomness $r_k$) let us assume the softmax form: $\prb(x) = S(\crb(x))$ where $\crb(x)$ is the classifier score vector of a base classifier and $S$ is the softmax function. Specialized to binary classification, $\prb(x) = (\prbp(x), 1-\prbp(x))$ where $\prbp(x) = s(\crb(x))$, $\crb(x)$ is a single real score and $s$ is the sigmoid function, $s(c) = 1/(1+\exp(-c))$. The key idea is to make a suitable assumption, for each $x$, on the distribution of either $\crb(x)$ or $\prb(x)$, over $r_k$; \textit{any suitable parametric distribution with two or three parameters can be used}. A good example is: {\it normality approximation} which says that, for each $x$, $\crb(x)$ is a normal random variable, $N(\mu(x),\sigma^2(x))$. This a decent assumption to make for stable classifiers such as regularized linear and kernel classifiers, k-nearest neighbor etc., as well as decision trees of limited size. Alternatively, $p_{r_k}^+(x)$ could be directly modeled using a beta distribution. The key idea that makes our approach efficient is that, given one of the distribution assumptions made above, we approximate $E_r L(p^*, p_r)$ of an ensemble using universal tables. For example, using the normality approximation, we approximate $E_r L(p^*, p_r)$ using a universal table that uses only $K$, $\mu$ and $\sigma^2$ as inputs. Together with the fact that $(\mu(x),\sigma^2(x))$ $\forall x$ can be estimated decently using a small set of (say 25) random base classifier instances of $r_k$ (Figure~\ref{fig:normality}), the whole process of computing $G$ $\forall K$ becomes very efficient. Demonstrating this clearly for binary classification is one of the other main aims of this paper. We point out how these ideas can be extended to multi-class problems using an additional approximation idea. 

We also consider an extra step - that of estimating suitably defined bias and variance components of $G$. While our idea of estimating $G$ applies to any loss $L$, for estimating bias-variance components we restrict ourselves to $L$ being either negative log-likelihood or least squares applied on the class probability vectors. For least squares loss in particular, we obtain closed form expressions that give useful insights. 


\section{Related work}
\label{sec:rw}
\begin{figure}[t]
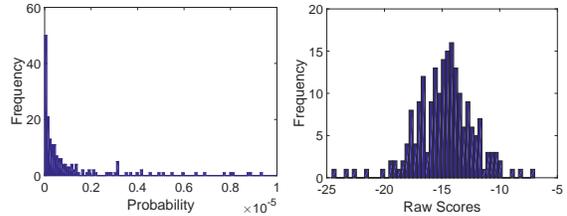

\centering
\includegraphics[width=0.45\linewidth]{uaiplots/probdistr_ijcnn.eps}
\includegraphics[width=0.45\linewidth]{uaiplots/normdistr_ijcnn.eps}
\caption{Distributions of classifier scores and probabilities for one example (from IJCNN Dataset), with base classifiers generated using the kernel method with polynomial kernel}
\label{fig:normality}
\end{figure}

We mainly review papers that contain ideas related to the estimation of errors of ensembles. Friedman~\cite{Friedman:97} approximates $p_r(x)$ by a normal distribution. In general, the distribution of $p_r(x)$ tends to be a skewed distribution especially when concentrated near $0$ or $1$ and hence, approximating $p_r(x)$ by a normal is poor; see figure~\ref{fig:normality} for an example. Also, this approximation allows values of $p_r$ outside the $(0,1)$ interval, creating issues for losses such as NLL. 


Schapire et al~\cite{Schapire:97} show that the 0/1 loss generalization error of voting methods is bounded by the number of training samples with a margin less than a threshold plus a term that depends on the number of training examples, the threshold, and the VC dimension of the base classifier. Zhang et al~\cite{Zhang:13} give a bound on the least squares error of parameter mixing applied to kernel ridge regression. Christmann et al~\cite{Christmann:07} give asymptotic bounds on the generalization error associated with an ensemble of kernel models  trained on small bites and use those bounds to promote the use of robust mixing for forming the ensemble. The bounds of~\cite{Schapire:97,Zhang:13,Christmann:07,zhou2012ensemble,galar2012review,ren2016ensemble} are all useful for obtaining various theoretical intuitions, but they are loose and not useful for the purpose of estimation. Hern\`{a}ndez-Lobato et al~\cite{Hern:07} give a monte-carlo scheme for estimating the generalization error of 0/1 loss for bagging. It is restricted to bagging. Also, their experiments show that it requires a large number of classifiers to be built in order to obtain estimates with decent quality. Mokeddem and Belbachir~\cite{Mokeddem:09} and Joao and Carlos~\cite{mendes2012ensemble} give a survey of a range of methods proposed in this area. These methods combine a variety of models with many classifier combination techniques and show the value of distributed ensembles using empirical analysis. There is also rich literature on bias-variance analysis and its use in explaining the working of ensemble methods such as bagging and boosting - see~\cite{Kuncheva:04}, ~\cite{Valentini:04}, ~\cite{buhlmann2012bagging}, ~\cite{kotsiantis2014bagging} and ~\cite{liu2017uncertainty} for good and compact summaries of the main ideas. Hence these works are orthogonal and complementary to our work in this paper; in particular, none of them give any scheme for estimating the generalization error (or its bias-variance components) of distributed ensemble.


\section{The main estimation ideas}
\label{sec:maincomp}

\def\Ltilde{\tilde{L}}

In this section, we explain our core ideas of approximating classifier score distribution for modeling randomness in bites selection and constructing universal table construction for voting and parameter mixing schemes. We present a procedure that can be used to estimate $G$ using the universal tables. We start with the definition of G that is appropriate for estimating the performance on real world datasets. 

\vspace*{-0.1in}
\subsection{Generalization error}
Let us consider the expectation of generalization error defined in (\ref{eq:morig}) and make an important practical change to it.
Almost all real world datasets do not give $p^*(x)$ for various $x$ values. Instead, they are given as a labeled set, $\{(x,y)\}$ where $x$ is an input and $y$ is a class label sampled from $p^*(x)$. In such situations, for doing practical computations, it is standard practice~\cite{Domingos:00,Valentini:04} to replace the generalization error $E_x L(p^*(x),p_r(x)$ with $E_{(x,y)} L(\delta(y),p_r(x))$ where $L$ is the loss function and $\delta(y)$ is a vector with all zeros except the component corresponding to $y$ which equals 1. Thus, for practical computations we take $G$ as
\begin{equation*}
G = E_{(x,y)} (E_r (L(\delta(y), p_r(x))))
\label{eq:mapprox}
\end{equation*}
Now let us look at binary classification for which $y\in\{1,0\}$. We can write $p_r(x) = (p_r^+(x), 1-p_r^+(x))$ where $p_r^+(x) = s(c_r^+(x))$ where $s$ is the sigmoid function and $c_r^+(x)$ is the positive class scoring function. 
Let us define $\Ltilde(y,p_r^+) = L(\delta(y), (p_r^+, 1-p_r^+))$ so that we can write $G$ as
\begin{equation*}
G = E_{(x,y)} (E_r (\Ltilde(y,p_r^+(x))))
\label{eq:revM}
\end{equation*}
In the following subsections, we explain how $E_r \Ltilde(y,p_r^+(x))$ can be computed efficiently for ensembles using parameter mixing and votes methods using score distributions and universal lookup tables. 

\vspace*{-0.1in}
\subsection{Approximation of score distribution}
First, we model the (pre-sigmoid) classifier score $c(x)$ generated via training on random small bites as normal for each $x$. Let $N(c(x)|\mu(x),\sigma^2(x))$, or, simply, $N(\mu(x),\sigma^2(x))$ denote this normal. We estimate the $\mu(x)$, $\sigma^2(x)$ $\forall x$ using just a few trained random base classifiers (25). We then use the normality approximation to estimate the generalization error of an ensemble efficiently. Note that our method is not particularly specialized for the normality approximation. Any suitable parametric distribution with two to three parameters can be used \footnote{See Appendix section A.3 for other possible distributions}. However, for stable classifiers such as regularized linear and kernel classifiers, k-nearest neighbor, etc., the normality approximation is a decent one to make. It is worth pointing out that, modeling the positive class probability (the output of the sigmoid) as a normal (Friedman, 1997) is not a good idea, especially when the probability approaches either 0 or 1. Figure~\ref{fig:normality} shows this clearly for one example from the IJCNN data for which a kernel method using polynomial kernel was used; see section~\ref{sec:expts} for details of the experimental setup. While the pre-sigmoid raw scores look normal, the probabilitiy values are asymmetrically distributed with a long tail on one side.  

\vspace*{-0.1in}
\subsection{Universal tables for mixing methods}
In this subsection, we derive expressions for the generalization error of two mixing schemes: (1) voting and (2) parameter mixing, and explain how the universal tables are constructed. In this paper we develop the ideas mainly for binary classification problems.

\subsubsection{Voting}
Voting consists of choosing the $y$ that is predicted by the most number of base classifiers. In the probabilistic version of voting we have $p_r^+ = mean \{s(c_k^+)\}$.
Now
\begin{equation*}
G = E_{(x,y)} (J_v(y,\mu(x),\sigma^2(x),K))
\label{eq:Mvint}
\end{equation*}
\begin{eqnarray*}
J_v(y,\mu(x),\sigma^2(x),K) = \\
\int \Ltilde(y,mean\{s(c_k^+)\}) \prod_k N(c_k^+|\mu(x),\sigma^2(x)) dc_k^+
\label{eq:Jvdef}
\end{eqnarray*}
and the subscript $v$ denotes voting. Let us define
\begin{equation*}
T_v(a,b,K) = \int \Ltilde(1,mean\{s(c_k^+)\}) \prod_k N(c_k^+|a,b) dc_k^+
\label{eq:uvint}
\end{equation*}
A universal table, $(a,b,K)\rightarrow T_v(a,b,K)$ can be built for a discrete choice of values of $(a,b,K)$ covering a range of interest. Then the value of $T_v(a,b,K)$ for any $(a,b,K)$ can be approximated by locating its position in the table and applying interpolation using the closest grid points. The key thing to note is that $T_v(a,b,K)$ is a universal table that does not depend on $x$ or the dataset; this is because $a$ and $b$ are scalar values on the real line and ($\mu(x)$, $\sigma^2(x))$ get mapped to closest scalar value entries for interpolation in the table. Therefore, the table just has to be built once for a given loss function. The multi-dimensional integral in the equation can be approximated simply by doing a joint sampling of the variables and then averaging the $\Ltilde$ values of the sample points. Because of the symmetry in the variables, the sample size needed for accurate estimation of the integral is not large; we found 5000 samples to be sufficient.
Now
\begin{equation*}
J_v(1,\mu(x),\sigma^2(x),K) = T_v(\mu(x),\sigma^2(x),K)
\label{eq:J1}
\end{equation*}
Let us also assume a symmetry property on the loss function:
$$\Ltilde(y,p_r^+) = \Ltilde(1-y,1-p_r^+)$$
Using this with the sigmoid property, $1-s(c) = s(-c)$,  we have
\begin{equation*}
J_v(0,\mu(x),\sigma^2(x),K) = T_v(-\mu(x),\sigma^2(x),K)
\label{eq:J0}
\end{equation*}
Thus, $G$ can be computed for $y\in\{0,1\}$ using the universal table $T_v$. If the symmetry condition on $\Ltilde$ doesn't hold, then we will have to build two universal tables - one for $y=1$ and one for $y=0$.

\subsubsection{Parameter mixing (PM)}
Here mixing is done by forming the mean of the classification scores of the base classifiers and then applying the sigmoid to obtain class probabilities~\cite{Mann:09,zinkevich:10}. For PM we have $p_r^+ = s(mean\{c_k^+\})$, where $\{c_k^+\}$ is the set of $K$ iid positive class-score random variables (each is $N(\mu(x),\sigma^2(x))$) associated with the $K$ base classifiers that make up the ensemble. Now
\begin{equation*}
G = E_{(x,y)} (J_{pm}(y,\mu(x),\sigma^2(x),K))
\label{eq:Mint}
\end{equation*}
\begin{eqnarray*}
J_{pm}(y,\mu(x),\sigma^2(x),K) = \\
\int \Ltilde(y,s(\sum_k c_k^+)) \prod_k N(c_k^+|\mu(x),\sigma^2(x)) dc_k^+
\label{eq:Jdef}
\end{eqnarray*}
Let us define
\begin{equation*}
T_{pm}(a,b,K) = \int \Ltilde(1,s(\sum_k c_k^+)) \prod_k N(c_k^+|a,b) dc_k^+
\label{eq:uint}
\end{equation*}
and compute $G$ as in voting, e.g., $$J_{pm}(1,\mu(x),\sigma^2(x),K)=T_{pm}(\mu(x),\sigma^2(x),K)$$.

In the case of PM, one can use the normality approximation to make a useful simplification. Because $c^+=mean\{c_k^+\}$ is $N(\mu(x),\sigma^2(x)/K)$, we have $T_{pm}(a,b,K) = \tilde{T}(a,b/K)$, where
\begin{equation*}
\tilde{T}(a,\tilde{b}) = \int \Ltilde(1,s(c^+)) N(c^+|a,\tilde{b}) dc^+
\label{eq:newuint}
\end{equation*}
Thus, it is sufficient to build a table with two parameters $(a,\tilde{b})$. Another comment that is worth making is that, when $K$ is large enough, then, even without the normality assumption, central limit theorem implies that $c^+$ is close to $N(\mu(x),\sigma^2(x)/K)$.

Algorithm~\ref{algo:algo1} summarizes the steps for estimating the generalization error for ensemble size $K$ using our approach.

\begin{algorithm2e}[h!]
\caption{Generalization Error Estimation using Universal Tables.}
\label{algo:algo1}
\SetKwInOut{Input}{Input}
\Input{Training Dataset $D$\\Test/Validation Dataset $V$\\Mixing Strategy - PM or Voting\\Corresponding pre-computed Universal Table,\\ $T(a,b,K)$ ((\ref{eq:uint}) or (\ref{eq:newuint}) for PM, (\ref{eq:uvint}) for voting)\\Number of base classifiers, $\Ktilde$ (a small number, say $25$)\\Size of ensemble, $K$ ($K > \Ktilde$)\\Bite size, $m$\\Base classifier Trainer, $T_c$. }
\KwResult{Generalization Error, $G$, on Test dataset, $V$ for ensemble size $K$.}
$G \longleftarrow 0$\;
Choose $\Ktilde$ random bites of size $m$ from $D$\;
Using $T_c$, train on the random bites to get $\Ktilde$ base classifiers, $\{c_{r_i}\}$\;
\ForEach{$(x,y) \in V$}{
Estimate mean ($\mu(x)$) and variance ($\sigma^2(x)$) of raw (pre-sigmoid) predictions $\{c_{r_i}(x)\}$\;
Find the generalization error for $x$ using the look-up table, $G_x \longleftarrow T((2y-1)\mu(x),\sigma^2(x),K)$\;
\tcp{Use bi-linear interpolaton since table is constructed for discrete values.}
$G \longleftarrow G + G_x$.
}
\end{algorithm2e}

\subsection{Estimating variance of the generalization error}
We have only talked about computing $G=E_{(x,y)} (E_r L(y,p_r(x)))$ till now. Let $G(x,y) = E_r( L(y,p_r(x)))$. We can apply the same estimation principles outlined above to compute $$S^2=E_{(x,y)} (E_r ((L(y,p_r(x))-G_{(x,y)}))^2)$$ Using the assumption that $L(y,p_r(x))$ over $(x,y)$ are independent and identically distributed, we can treat this computation as an approximation of the variance \footnote{This is the variance of the generalization error. It should not be confused with the variance component of $G$, which we analyze in section~\ref{sec:bv}.} 
of $E_{(x,y)} (L(y,p_r(x)))$. We can compute $S^2$ as $$S^2 =  E_{(x,y)} (E_r (L^2(y,p_r(x)))) - E_{(x,y)} (G^2(x,y)).$$

\def\jtilde{{i}}

\section{Bias-Variance estimation}
\label{sec:bv}

In this section we show how the estimation principles developed in section~\ref{sec:maincomp} can be applied to the estimation of bias-variance components of $G$. We begin by describing the basic definitions and theory and then develop the estimators.

\def\pbar{\bar{p}}

\subsection{Basic theory}
Let us write the ideas for a general classifier - it could be a single base classifier or an ensemble.
To make the ideas clearly visible, let us first fix an $x$ and also not mention any dependence on $x$. 
The ideas developed here apply to a special class of loss functions $L(p^*,p_r)$ in which the interaction terms involving $p^*$ and $p_r$ appear linearly in $p^*$. But, here we get specific and describe the ideas only for two important losses that belong to this class - negative log-likelihood (NLL) and least squares (LS):
\begin{eqnarray*}
NLL: \; L(p^*,p_r) = -p^*\cdot \log(p_r)
\end{eqnarray*}
\begin{eqnarray*}
LS: \; L(p^*,p_r) & = & \|p^*-p_r\|^2 \\
                  & = & \|p^*\|^2 - 2p^*\cdot p_r + \|p_r\|^2
\label{eq:1}
\end{eqnarray*}
Let $\pbar$ be the mean distribution that is obtained as
\begin{equation*}
\pbar = \arg \min_q E_r\, D(q,p_r)
\label{eq:2}
\end{equation*}
where $D$ is a suitable distance metric between two class distributions. Negative Log Likelihood(NLL) uses Kullback-Leibler divergence~\cite{Heskes:98} and least square (LS) uses Euclidean distance~\cite{Friedman:97}. This yields
\begin{eqnarray*}
NLL: \;\; \pbar = \exp (E_r \log p_r)/Z, \; Z=e\cdot E_r\log p_r, \\
LS: \;\; \pbar = E_r p_r
\label{eq:3}
\end{eqnarray*}
where $e$ is a vector of all ones. Using these it is easy to show~\cite{Heskes:98} that 
\begin{eqnarray*}
NLL: \;\; E_r L(p^*,p_r) = L(p^*,\pbar) - \log Z, \\
LS: \;\; E_r L(p^*,p_r) = L(p^*,\pbar) + e\cdot var(p_r)
\label{eq:4}
\end{eqnarray*}
where $var(p_r)$ is a vector containing the variances of the components of $p_r$. Thus, the full generalization error's bias-variance decomposition is given by
\begin{equation*}
E_x E_r L(p^*,p_r) = B + V, \;\;\;\; B = E_x L(p^*,\pbar),
\label{eq:5}
\end{equation*}
\begin{eqnarray*}
NLL: \;\; V = E_x -\log Z \;\;\;\; LS:  \;\; V = E_x e\cdot var(p_r)
\label{eq:6}
\end{eqnarray*}

\subsection{Estimation of bias-variance components for ensemble} 
For practical computation with training sets which only have samples of $(x,y)$, we replace $B$ by $B = E_{x,y} L(\delta(y),\pbar)$. It is clear that, for computing $B$ and $V$, we need an efficient way of computing $E_r \log p_r(x)$ and $E_r p_r(x)$ for each $x$. For binary classification, it is sufficient to compute $E_r \log p_r^+(x)$ and $E_r p_r^+(x)$ for each $x$. For parameter mixing and voting, these quantities can be estimated using tables, using ideas similar to those discussed in section~\ref{sec:maincomp} for $\Ltilde(y,p_r)$ \footnote{See appendix section A.2 for more details}.

\section{Experiments}
\label{sec:expts}
In this section we conduct several experiments to cover various aspects of the ensemble design problem, as listed below. 
\begin{itemize}
\item We demonstrate that the accuracy of our analytical estimators is consistently better than a practical baseline on several datasets as measured by the (relative) generalization error (0/1 loss) with respect to the ground truth. We observe a similar consistent behavior for the NLL bias and variance estimators as well. 
\item Through real world examples, we illustrate how the optimal number of classifiers and bite size can be determined using our analytical estimators. We also discuss empirical results on the choice of parameter mixing and voting scheme under different bite size and number of classifier model scenarios.
\end{itemize}


\subsection{Data}
We conducted experiments on three datasets, IJCNN, LETTER and KDD~\cite{libsvm:16}. All estimations were done on the test set. For LETTER, we divided the $26$ letters into two 
classes pairs to create a binary classification problem.  It is worth pointing out 
that our experiments cover a variety of models: linear classifiers, kernel methods and decision trees. 
We used LIBSVM~\cite{chang:11} to train the kernel classifiers. Details associated with the datasets and the experiments are given in  Table~\ref{tab:details}.

\begin{table}[ht]
\vspace*{-0.1in}
\small
\caption{Details of datasets, base classifiers and other experimental parameters. Here K is the number of base classifiers forming an ensemble, m is the size of small bites of training data, and  $extclass$ is the number of base classifiers used for estimating $~\mu$ and $~\sigma^2$.}
\vspace*{-0.1in}
\label{tab:details}
\begin{center}
\begin{tabular}{|c|c|c|c|}
\hline
Dataset & \# training           & \# test            & Base classifier     \\
        & examples ($n$)        &  examples          &                     \\
\hline
KDD     & $8.4\times 10^6$      & $5.1\times 10^5$   & Linear SVM          \\
\hline
IJCNN   & $5\times 10^4$             & $9.2\times 10^4$               & Poly SVM \\
\hline
LETTER  & $1.5\times 10^4$             & $5\times 10^3$               & Decision Trees     \\
\hline
\end{tabular}

\begin{tabular}{|c|c|c|c|}
\multicolumn{4}{}{} \\
\hline
Dataset  & $K$     &   $m/n$       & $extclass$ \\
\hline
KDD      & 1-200   &  0.01, 0.02, 0.04, 0.1    & 25         \\
\hline
\multirow{ 2}{*}{IJCNN}    & \multirow{ 2}{*}{1-800}   & 0.01, 0.02, 0.04,  & \multirow{ 2}{*}{25}  \\
& & 0.1, 0.25, 1.0 &\\
\hline
\multirow{ 2}{*}{LETTER}   & \multirow{ 2}{*}{1-800}   & 0.01, 0.02, 0.04,  & \multirow{ 2}{*}{25}  \\
& & 0.1, 0.25, 1.0 &\\
\hline
\end{tabular}
\end{center}
\end{table}



\def\nbmusig{\Ktilde}

\subsection{Samples For $(\mu,\sigma^2)$ Estimation}
\begin{figure}[h]
\centering
\includegraphics[width=0.7\linewidth]{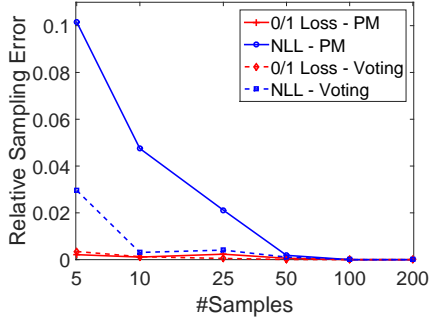}
\caption{IJCNN: \# samples for $(\mu,\sigma^2)$}
\label{fig:sampleeffect}
\end{figure}

Let us begin by making a note on $\nbmusig$, the number of base classifiers used for estimating $\mu$ and $\sigma^2$. We found that a small value of $\nbmusig$ is sufficient to obtain good and useful estimates of $\mu$ and $\sigma^2$ needed by our analytical method . All results
of this section are presented for estimators built with $\nbmusig=25$. Figure~\ref{fig:sampleeffect} clearly illustrates this for one instance  of the IJCNN dataset.

\subsection{Baselines}
We consider the following three intuitive baselines for the qualitative and quantitative comparison of our approach. The methods are presented in the order of decreasing estimation accuracy, which also turns out to be in the order of decreasing computational cost.


{\bf{Empirical Estimation or Ground Truth ({\it{GT}}): }} In order to estimate the ground truth empirically,  we form a collection of 
2000 random base classifiers for each $m$\footnote{One may build a lot more than 2000 base classifiers, to make a better ground truth. For the three chosen datasets we found 2000 to be a sufficient number}.
For each given $K$, a large number of ensembles 
of size $K$ were formed using the above collection. 
Expectation was done using these ensembles 
to obtain the ground truth estimate of the expectation of any quantity. Note that since this method requires a lot of classifiers to perform the expectation correctly, it is very costly in practice.

\begin{figure*}[t]
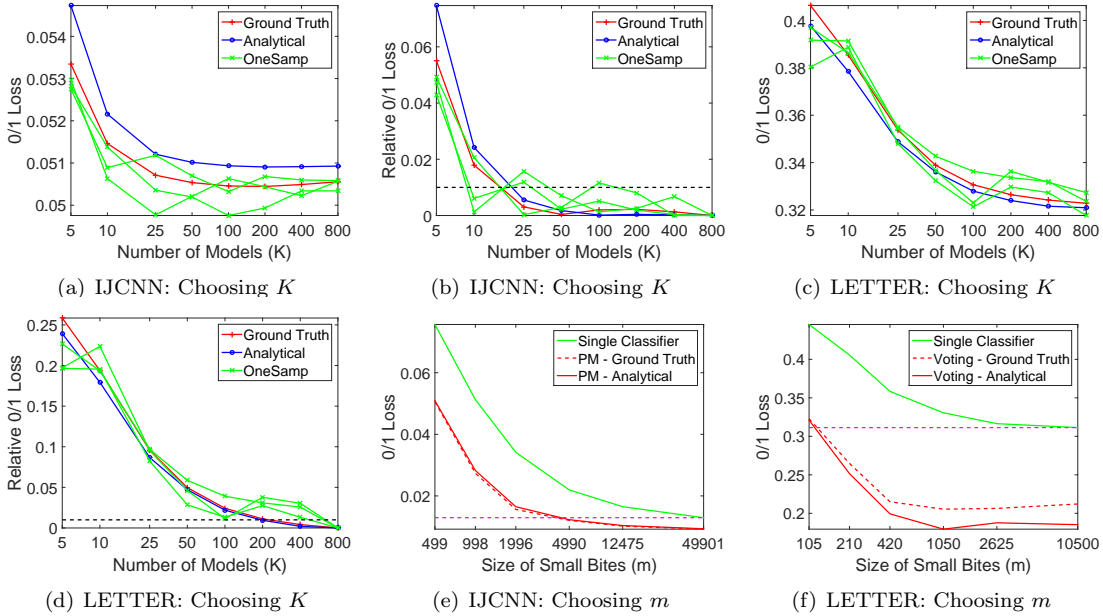

\centering
\subfigure[IJCNN: Choosing $K$]{
\includegraphics[width=0.28\linewidth]{uaiplots/pm_nummodels_ijcnn.eps}
\vspace*{-2in}
}
\subfigure[IJCNN: Choosing $K$]{
\includegraphics[width=0.28\linewidth]{uaiplots/pm_nummodels_rel_ijcnn.eps}
\vspace*{-2in}
}
\subfigure[LETTER: Choosing $K$]{
\includegraphics[width=0.28\linewidth]{uaiplots/ens_nummodels_letter.eps}
\vspace*{-2in}
}
\subfigure[LETTER: Choosing $K$]{
\includegraphics[width=0.28\linewidth]{uaiplots/ens_nummodels_rel_letter.eps}
\vspace*{-2in}
}
\subfigure[IJCNN: Choosing $m$]{
\includegraphics[width=0.28\linewidth]{uaiplots/ijcnn_pm_lc.eps}
\vspace*{-2in}
}
\subfigure[LETTER: Choosing $m$ ]{
\includegraphics[width=0.28\linewidth]{uaiplots/letter_ens_lc.eps}
\vspace*{-2in}
}
\caption{Value of our method for choosing $K$ and $m$. Plots (a)-(d) concern the choosing of $K$; (a) and (c) show absolute estimation values of 0/1 loss, while (b) and (d) show relative values normalized by each curve's value at $K=800$. In (b) and (d) the black dotted line shows the relative error of 1\%. Plots (e) and (f) concern the choosing of $m$. The magenta dotted line shows the error of single classifier trained on full data.}
\label{fig:others}
\vspace*{-0.1in}
\end{figure*}

{\bf{One Sample Estimation ({\it{OneSamp}}): }} A practical and intuitive approach is to keep on building base classifiers incrementally until adding more base classifiers does not improve the performance 
significantly. We call this as {\em one sample estimation} since we are just using one sample for any given value of $K$ to calculate the quantities. There are two potential issues with this incremental approach: a) the estimates will be jumpy and unstable if the variance is very high (see Figures~\ref{fig:others}(a)-(d)). This introduces significant errors in the estimation of optimal ensemble size, and, b) the degree of parallelization in building ensembles is limited in the distributed setting since checking over test data needs to be done after adding few models.

{\bf{Use only $\nbmusig$ models for ensembles ({\it{EmpSamp-$\nbmusig$}}): }} Another possible approach is to use only $\nbmusig$ models (i.e. $25$ in our setting) and do the empirical estimation using only them. 
Note that for $K<\nbmusig$, we can form random subsets of $K$ classifiers to perform the expectation. The obvious issue with this approach is that in many cases (especially for big datasets) the number of models required for good performance are significantly larger than $\nbmusig$ (Figure~\ref{fig:others}(a)-~\ref{fig:others}(d)). As a result, this approach has limited practical utility and will not be considered further.

\begin{figure*}[h!]
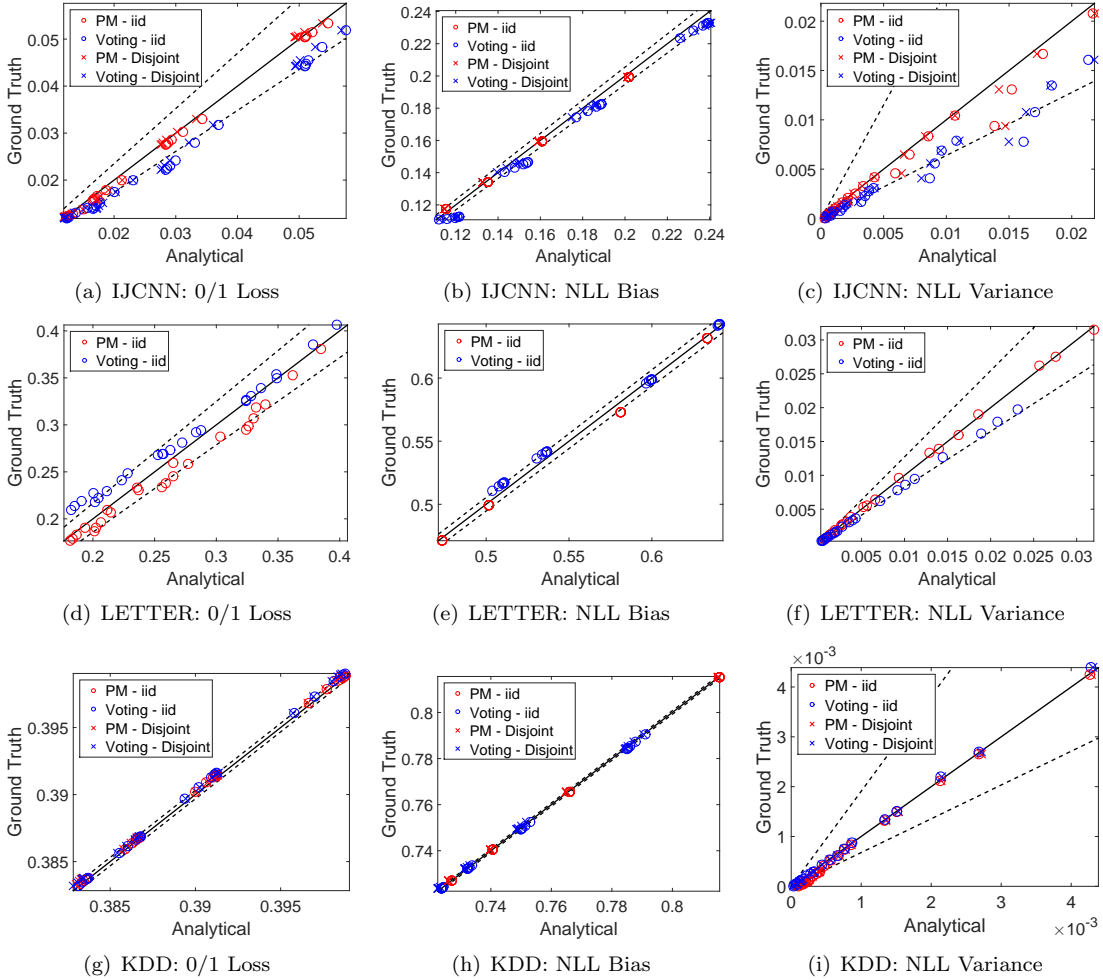

\centering
\subfigure[IJCNN: 0/1 Loss]{
\includegraphics[width=0.28\linewidth]{uaiplots/ijcnnscatterplot01.eps}
\vspace*{-2in}
}
\subfigure[IJCNN: NLL Bias]{
\includegraphics[width=0.28\linewidth]{uaiplots/ijcnnscatterplotbiasnll.eps}
\vspace*{-2in}
}
\subfigure[IJCNN: NLL Variance]{
\includegraphics[width=0.28\linewidth]{uaiplots/ijcnnscatterplotvarnll.eps}
\vspace*{-2in}
}
\subfigure[LETTER: 0/1 Loss]{
\includegraphics[width=0.28\linewidth]{uaiplots/letterscatterplot01.eps}
\vspace*{-2in}
}
\subfigure[LETTER: NLL Bias]{
\includegraphics[width=0.28\linewidth]{uaiplots/letterscatterplotbiasnll.eps}
\vspace*{-2in}
}
\subfigure[LETTER: NLL Variance]{
\includegraphics[width=0.28\linewidth]{uaiplots/letterscatterplotvarnll.eps}
\vspace*{-2in}
}
\subfigure[KDD: 0/1 Loss]{
\includegraphics[width=0.28\linewidth]{uaiplots/kddscatterplot01.eps}
\vspace*{-2in}
}
\subfigure[KDD: NLL Bias]{
\includegraphics[width=0.28\linewidth]{uaiplots/kddscatterplotbiasnll.eps}
\vspace*{-2in}
}
\subfigure[KDD: NLL Variance]{
\includegraphics[width=0.28\linewidth]{uaiplots/kddscatterplotvarnll.eps}
\vspace*{-2in}
}
\caption{Scatter plots comparing analytical estimates against ground truth, for IJCNN, LETTER and KDD. The dotted lines represent the 3/4 quantile of relative error of the analytical estimate. The relative error (which corresponds to the slope in the figure)
values associated with the quantiles are as follows. IJCNN: 0/1 loss: $0.148$, NLL bias: $0.026$, NLL variance: $0.565$; LETTER: 0/1 loss: $0.076$, NLL bias: $0.011$, NLL variance: $0.214$ and KDD: 0/1 loss: $0.81\times 10^{-3}$, NLL bias: $0.98\times 10^{-3}$, NLL variance: $0.476$. The slopes of the dotted lines have values $1\pm$ relative error.}
\label{fig:scatter}
\end{figure*}

\subsection{Estimation Errors}
We now do the quantitative evaluation of our analytical estimates. There are many dimensions to study: $m$, $K$, PM/voting, and whether small bites were generated with replacement (iid) or by partitioning the full data (Disjoint). Doing a detailed study of relative errors covering all dimensions is cumbersome. So we restrict ourselves to a simplified study. Figure~\ref{fig:scatter} compares, for IJCNN, LETTER and KDD, analytically estimates against ground truth ({\it{GT}}) via scatter plots, separately for 0/1 loss, NLL bias and NLL variance. The points in the scatter plots are obtained by sweeping over various values of the dimensions. The dimensions, PM/voting and iid/disjoint\footnote{For LETTER, we show results with iid sampling only, since we used an external package (proprietary) for building decision tree ensemble which does not 
support disjoint sampling.} are differentiated within each scatter plot using different colors.
The dimensions, $m$ and $K$ did not show any interesting patterns of relative errors in estimation. The dotted lines represent the 
3/4 quantile of relative error of the analytical estimate, i.e., $75\%$ of the points lie within the dotted lines.


We can make the following observations based on figure~\ref{fig:scatter}. (a) The analytical estimates are better for PM than for voting. (b) Relative errors are somewhat large for NLL variance. However, note that, for all datasets, the higher error occurs when the variance component is very small and the bias component dominates the variance; hence, even large relative errors in variance (e.g., 0.5-0.75) are inconsequential as far as the total NLL estimation is concerned. (c) This paper does not give a method for computing bias and variance components of ensemble classifiers for 0/1 loss. If one is interested in computing them, they have to resort to the fundamental ideas given by (Domingos, 2000) and employ methods such as the direct sampling method outlined in section~\ref{sec:intro}; see also (Valentini et al, 2004). (d) There is no noticeable difference between iid and disjoint methods of forming base classifiers. However we observe that the relative errors for KDD are much smaller than those for IJCNN and LETTER. 

\vspace*{-0.1in}
 \subsection{Usefulness in Design - Optimal Models ($K$)}
Figures~\ref{fig:others}(a)-~\ref{fig:others}(d) evaluate our method to choose $K$ for the IJCNN and LETTER datasets with  $m/n = 0.01$.  Three independent runs of {\it{OneSamp}} baseline are also shown in the plot. Figures~\ref{fig:others}(a) and~\ref{fig:others}(c) show the 0/1 loss as a function of the number  of models ($K$). Note that for both the datasets our analytical estimates are very close to the ground truth\footnote{Visually, the differences between our estimates and ground truth may look big for IJCNN (Figure~\ref{fig:others}a). However, note that the x-axis starts with $K=5$ and the range 
of y-axis is very small.} On the other hand, {\it{OneSamp}} estimates are jumpy and the three runs show a lot of variance. Figures~\ref{fig:others}(b) and~\ref{fig:others}(d) show the corresponding relative 0/1 loss plots with respect to the loss at $K=800$. Let us say we want to choose $K$ to be the smallest value such that the 0/1 loss reaches within 1\% of its asymptotic value (dotted black line) as $K\to\infty$. For LETTER, the ground truth evaluation led to a value of $K=271$. Our method gave an estimate of $K=246$, which is very close to the true estimate. On the other hand, none of the {\it{OneSamp}} baseline curves go below the 1\% line except at $K=800$ since it is the reference point for relative 0/1 loss computations. For IJCNN, our analytical estimate predicted $K=21$ wile ground truth evaluation was $K=18$. On the other hand the three {\it{OneSamp}} estimations were $10$, $10$ and $25$, showing a significant variance. 
\vspace*{-0.1in}
\subsection{Usefulness in Design - Optimal Bite size ($m$)}
Figures~\ref{fig:others}(e) and (f) illustrate the use of our analytical method for choosing $m$ for the PM ensemble method for IJCNN and Voting ensemble method for LETTER respectively. Suppose we are interested in determining the least value of $m$ such that the ensemble method achieves the same expected 0/1 loss as the expected loss of the single classifier using the full data. For IJCNN, the use of ground truth gave a value of $m=3539$ while the analytical method led to the estimate, $m=3624$. For the voting method, the corresponding values were $m=2640$ and $3640$; the large difference in values is due to the larger relative errors of our method for voting. For LETTER, the ground truth gave value of $m=120$ while the analytical method predicted $m=115$ for voting. 



\vspace*{-0.1in}
\subsection{Usefullness in Design - PM vs. Voting}
On the IJCNN dataset, we varied $m/n$ (small bites as a fraction of the full data) and $K$, and, for each situation applied our estimators of NLL (the sum of bias and variance components) to determine whether PM or voting had better generalization error. On all cases except $m/n=0.1$, the estimators gave the correct decision. For  $m/n=0.1$, we found the PM and voting estimates to be close to each other (within 3\% (relative)). Hence, our analytical estimates could not predict correctly and if estimates are so close, it does not really matter to choose PM or voting.


\section{Conclusion and Discussion}
\label{sec:conc}
In this paper we have proposed an efficient method for estimating the expected generalization error and its bias-variance components. There are many directions for improvement and future work. (1) Since we use only a small number of base classifiers, would Bayesian modeling, as shown in ~\cite{gosink2017bayesian} lead to a more effective determination of the $\mu(x), \sigma^2(x)$ parameters? (2) extension to multi-class problems is an important direction \footnote{See appendix section A.1 for more details}. (3) Trying out new distributions other than normal and beta is another worthy direction \footnote{See appendix section A.3 for more details}. (4) It would be useful to apply the ideas to the design of bagging classifiers, random forests and building distributed ensembles \footnote{See appendix section A.4 for more details}. (4) For 0/1 loss, since the modeling around the zero classification score is more important, higher weighting of samples with scores close to zero in the estimation of $\mu(x), \sigma^2(x)$ could lead to better estimates of $G$. We have done initial experiments to verify this, but more work is needed. 

\appendix
\section{Appendix}
\subsection{Extension to multi-class}
In multi-class problems $c(x)$ has many components. Even if we make the reasonable approximation that the components can be modeled as independent normal random variables, they affect $p_r$ in a nonlinear fashion via softmax, requiring the evaluation of integrals that are joint functions of all $(\mu^j,(\sigma^j)^2)$ pairs associated with the components of $c(x)$. One approach to first write each component, $p_r^j$ as
\begin{eqnarray*}
p_r^j = \exp(c^j)/\sum_\jtilde \exp(c^\jtilde) = s(\alpha), \\
\alpha = \log \sum_{\jtilde\not= j} \exp(c^\jtilde-c^j)
\label{eq:mclass1}
\end{eqnarray*}
We can linearize $\alpha$ around the $(\mu^j,(\sigma^j)^2)$ pairs to write it in the linear form, $\alpha\approx \beta^0 + \sum_\jtilde \beta^\jtilde c^\jtilde$. With this approximation, $\alpha$ is a normal random variable $N(\alpha | \beta^0 + \sum_\jtilde \beta^\jtilde\mu^\jtilde, \sum_\jtilde (\beta^\jtilde\sigma^\jtilde)^2)$. Then the estimation computations can proceed as in the binary case. It is easy to see that, for the binary classification case, this approximation is exact because $\alpha$ itself is linear. The above ideas for multi-class problems are yet to be tested.


\subsection{Closed form expressions for LS.}
Recall that, for LS, we have, for any classifier with randomness $r$,
\begin{eqnarray*}
B=E_xB(x), \; B(x) =  \|p^*(x)-\pbar(x)\|^2, \; V = E_x V(x)
\end{eqnarray*}
\begin{eqnarray*}
V(x) =  e\cdot var(p_r(x)), \;\; \pbar(x) = E_r p_r(x)
\label{lsrecall}
\end{eqnarray*}
As before, let us fix one $x$ and leave out the reference to $x$ at most places to avoid clumsiness.
Let us use subscripts $b$, $v$ and $pm$ to denote, respectively, single base classifier, voting and parameter mixing.\vspace*{0.05in}

\subsubsection{Single base classifier}
For binary classification we derive approximate expressions for $B_b(x)$ and $V_b(x)$.
\begin{equation*}
\pbar_b^+ = E_r p^+ = \int s(c) N(c|\mu,\sigma^2) dc
\label{eq:empbar}
\end{equation*}
Using results from~\cite{Mahajan:12} - see section 6 there and also Exercise 4.26 in~\cite{Bishop:06} - we can derive the approximation
\begin{eqnarray*}
\pbar_b^+ = s(\kappa(\sigma^2)\mu), \;\;\; \kappa(\sigma^2)  = (1 + \pi \sigma^2/8)^{-1/2}
\label{eq:pbarexp}
\end{eqnarray*}
\begin{eqnarray*}
B_v(x) = & B_b(x) = & 2(\pbar^+ - p^{*+})^2  \\ = 2(s(\kappa(\sigma^2)\mu) - p^{*+})^2
\label{eq:Bexp}
\end{eqnarray*}
and also the variance approximation
\begin{eqnarray*}
V_b(x) = 2[s^2(\rho(\sigma^2)\mu + \delta [1-\rho(\sigma^2)]) - s^2(\kappa(\sigma^2)\mu)], \\
\;\;\; \rho(\sigma^2) = (1 + \lambda^2\sigma^2)^{-1/2}, \;\;\; \lambda=0.703\;\;\; , \;\;\; \delta=0.937
\label{eq:varexp}
\end{eqnarray*}

\subsubsection{Ensemble with voting and parameter mixing}
We give the bias and variance results for the voting scheme when we use $K$ base classifiers. 
\begin{eqnarray*}
\pbar_v  = E_r p_r  = mean \{E_{r_k} p_{r_k} \} = \pbar_b, \;\;\;\; B_v(x) = B_b(x)
\label{eq:pbarv}
\end{eqnarray*}
\vspace*{-0.3in}
\begin{eqnarray*}
V_v(x) = e\cdot var(p_r),
\end{eqnarray*}
\vspace*{-0.3in}
\begin{eqnarray*}
var(p_r)  =  E_{\{r_k\}} \| mean\{p_{r_k}\}-\pbar_b \|^2 
                                         =  \frac{1}{K} V_b(x)
\end{eqnarray*}
It is useful to study the pros and cons associated with the variance in scores. Note that, as $\sigma^2\to 0$, both $\kappa(\sigma^2)$ and $\rho(\sigma^2)$ increase towards 1. Therefore $\pbar^+$ moves from $1/2$ (when $\sigma^2$ is large) to $s(\mu)$ as $\sigma^2\to 0$.
The behavior of bias depends on the location of $p^{*+}$ with respect to $s(\mu)$. If both of these are on the same side of $1/2$, i.e., both $p^{*+}$ and $s(\mu)$ are greater than $1/2$, or, both are smaller than $1/2$, we call that as the {\em Unbiased case}; else, we call that as the {\em Biased case}. For $p^{*+}$ taking extreme values, i.e., $0$ or $1$ it is easy to see that (a) for the unbiased case, bias decreases as $\sigma^2$ decreases; and (b) for the biased case, bias increases as $\sigma^2$ decreases. These observations are similar to what are observed by (Domingos, 2000) and (Valentini and Dietterich, 2004), but here they come out a lot more concretely through expressions.\vspace*{0.05in}

For  parameter mixing, $c = \frac{1}{K} \sum_k c_k$. Since $\{c_k\}$ are iid and each $c_k$ is $N(\mu,\sigma^2)$, $c$ is normal and more specifically, it is $N(\mu,\sigma^2/K)$. Thus the expressions for $V_{pm}(x)$ and $B_{pm}(x)$ are same as that of $V_{b}(x)$ and $B_{b}(x)$, with $\sigma^2$ replaced by $\sigma^2/K$.
To compare a single classifier and PM, all that we again need to do is to study what happens to $V_b(x)$ and $B_b(x)$ as $\sigma^2\to 0$.

\subsection {A general view of the approach}

It is easy to note that, the basic idea behind our approach is not crucially dependent on the normality assumption or the link function being the sigmoid. \textit{Actual quantification of the goodness of this approximation is not so crucial. More important is the estimation error it leads to}. It allows more general modeling with other types of distributions and link functions. Let us give a few examples as potential possibilities. (a) We can move away from probabilistic modeling and replace the sigmoid by other functions, e.g., the unit step function to directly model 0/1 and other non-probabilistic losses. In fact this is the approach we took for all experiments related to 0/1 loss in section~\ref{sec:expts}. (b) We can directly approximate $p_r$ using a beta distribution. We have experimented with this for decision trees and obtained good results. But we do not give details due to lack of space.(c) We can model the score as a mixture of two normals (or using Kernel Density Estimation). This leads to an extra 3 parameters; forming the needed universal tables for this is more expensive to compute and store, but is manageable.This allows our approach to be applied to models such as decision trees which do not go through a score and a link function to form class probabilities.

\subsection{Applicability to Big Data}
 We discuss briefly how our analysis can be applied to the  data scenario on distributed platforms like Hadoop MapReduce~\cite{Chu:2006}, MPI~\cite{MPI:1994}, etc. We assume that each node gets a disjoint data partition/bite instead of iid samples. Note that there is no much difference between these two possibilities when the data size is large. 

One primary advantage of our analytical approach is that it is resource efficient in a distributed setting than the {\it{GT}} baseline\footnote{We will not consider other baselines since they are either unstable or of limited practical utility.}. Consider the task of finding the optimal ensemble size for given bite size $m$. {\it{GT}} will first train large number ($2000$ in our case) of models in a distributed fashion. Note that the disjoint setting might not even be able to generate $2000$ bites of size $m$ and, we will need to do costly data shuffle to generate the data for training. It will then communicate all the models to a central place where a validation set will be used to determine the optimal size, as discussed in previous section. Alternatively, each node will compute predictions on validation set for classifiers it has trained and communicate them to one central node. Nevertheless, in addition to using large amounts of cluster resources for training classifiers in the beginning, there is also a big communication overhead. Then, in the final step, making different combinations of classifiers and averaging their results also takes a lot of time. 

On the other hand, our analytical approach needs to train only $\Ktilde$ base models in the beginning and as discussed in the previous section, around $25-50$ are usually enough. As a result, both training (total machine hours) and communication times are $2$ orders of magnitude smaller. Moreover, table look-ups at central node are extremely fast. Subsequently, once optimal ensemble size $K$ is determined, $K$ models are trained in parallel. Typically, $K$ is going to be significantly smaller than the number of models trained by the baseline, thus giving our method a significant advantage. Our method can also efficiently find optimal bite size in a distributed setting. Since we need to train only a small number of base models in the beginning, we can assign a sufficiently large bite size $m_{max}$ to each partition only once, and then train base classifiers for different bite sizes $m < m_{max}$. This saves a considerable amount of remote data IO and job set-up times. Once, we have determined the optimal bite and ensemble size, we run the training again for only this setting.

\vspace*{-0.1in}
\bibliographystyle{abbrv}
\bibliography{cikm_ensembles}

\end{document}